%% file: main.tex
\documentclass[letterpaper]{article}

\usepackage{booktabs}
\usepackage{wrapfig}
\usepackage{float}

\usepackage{amssymb}
\usepackage{enumitem}
\usepackage{amsmath}
\usepackage[table, dvipsnames]{xcolor}

\usepackage[linesnumbered,lined,commentsnumbered,noend]{algorithm2e}
\usepackage{tcolorbox}                 

\makeatletter
\newcommand{\removelatexerror}{\let\@latex@error\@gobble}
\makeatother

\definecolor{seed_light}{RGB}{233,239,236}
\definecolor{seed}{RGB}{0,128,128}
\definecolor{nov}{RGB}{0,102,204}
\definecolor{exp}{RGB}{180,0,0}
\newcommand{\ColorLine}[2]{%
  {%
    \begingroup
      \SetNlSty{}{\color{#1}}{}%
      \textcolor{#1}{#2}%
    \endgroup
  }%
}

\newcommand{\Nov}[1] {\ColorLine{nov}{#1}}
\newcommand{\Exp}[1] {\ColorLine{exp}{#1}}
\makeatother
\makeatletter
\renewcommand{\@algocf@capt@plain}{above}
\makeatother

\SetKwFor{For}{for}{}{}               
\SetKwFor{uFor}{for}{}{endfor}              
\SetKwFor{While}{while}{}{endwhile}         

\DeclareMathOperator*{\argmin}{arg\,min} 

\usepackage{natbib,hyperref,alifeconf}  
\usepackage{url,cleveref}
\definecolor{myblue}{rgb}{0.25,0.25,0.55}
\definecolor{myred}{rgb}{0.8,0,0}
\hypersetup{
    colorlinks,
    linkcolor={myblue},
    citecolor={myblue},
    urlcolor={myblue},
    pdfproducer={},
    }
\newcommand{\todo}[1]{}
\renewcommand{\todo}[1]{{\color{myred} Todo: {#1}}}

\title{Expedition \& Expansion: Leveraging Semantic Representations for\\Goal-Directed Exploration in Continuous Cellular Automata}

\author{
    Sina Khajehabdollahi$^{1}$,
    Gautier Hamon$^{1}$,
    Marko Cvjetko$^{1}$,\\
    \Large
    Pierre-Yves Oudeyer$^{1}$,
    Cl\'ement Moulin-Frier$^{1}$,
    C\'edric Colas$^{1, 2}$\\
    \mbox{}\\
    $^1$Flowers AI \& CogSci Lab, Inria, France \\
    $^2$MIT, USA \\
    \texttt{sina.abdollahi@gmail.com}
} 

\begin{document}

\maketitle

\begin{abstract}
Discovering diverse visual patterns in continuous cellular automata (CA) is challenging due to the vastness and redundancy of high-dimensional behavioral spaces. Traditional exploration methods like Novelty Search (NS) expand locally by mutating known novel solutions but often plateau when local novelty is exhausted, failing to reach distant, unexplored regions. We introduce \textit{Expedition \& Expansion} (E\&E), a hybrid strategy where exploration alternates between \textit{local novelty-driven expansions} and \textit{goal-directed expeditions}. During expeditions, E\&E leverages a Vision-Language Model (VLM) to generate linguistic goals—descriptions of interesting but hypothetical patterns that drive exploration toward uncharted regions. By operating in semantic spaces that align with human perception, E\&E both evaluates novelty and generates goals in conceptually meaningful ways, enhancing the interpretability and relevance of discovered behaviors. Tested on Flow Lenia, a continuous CA known for its rich, emergent behaviors, E\&E consistently uncovers more diverse solutions than existing exploration methods. A genealogical analysis further reveals that solutions originating from expeditions disproportionately influence long-term exploration, unlocking new behavioral niches that serve as stepping stones for subsequent search. These findings highlight E\&E's capacity to break through local novelty boundaries and explore behavioral landscapes in human-aligned, interpretable ways, offering a promising template for open-ended exploration in artificial life and beyond.
\end{abstract}

\noindent
 Visualization/Code: \\ \href{https://developmentalsystems.org/expedition_expansion/}{developmentalsystems.org/expedition\textunderscore expansion}

\section{Introduction}
\begin{figure*}[t]
    \centering
    \includegraphics[width=0.9\linewidth]{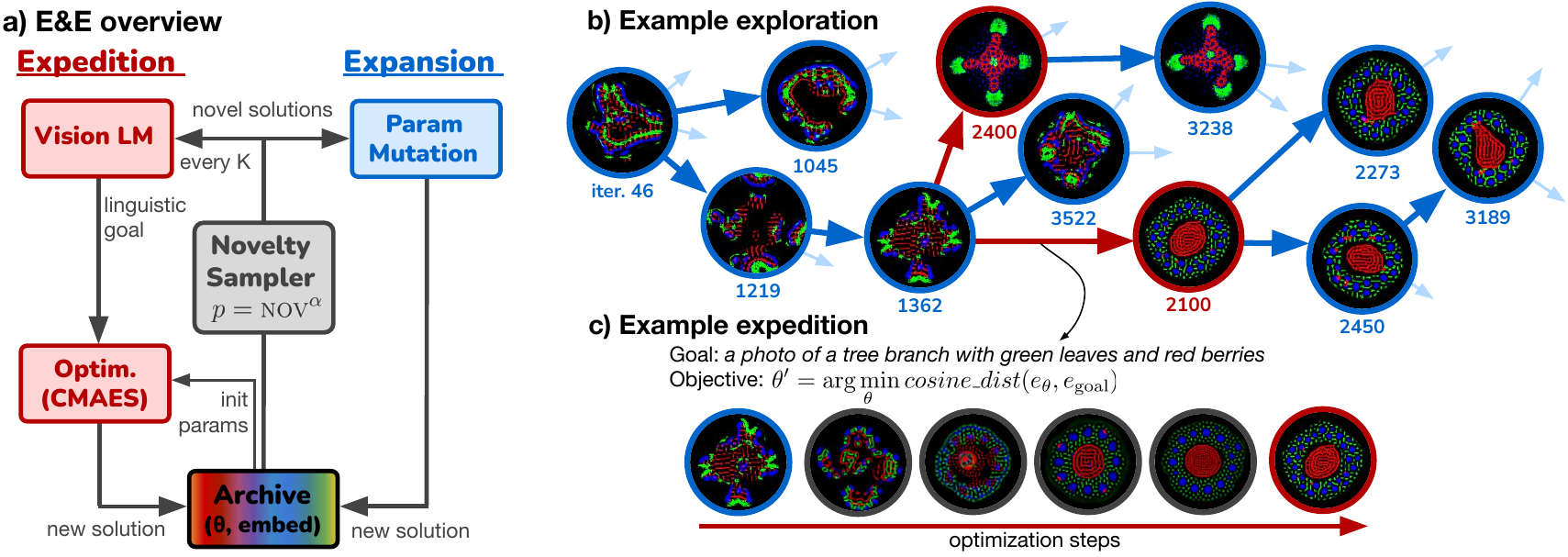}
    \caption{\textbf{Overview of \textit{Expedition \& Expansion}.} 
    (a) E\&E alternates between expansion steps guided by Novelty Search and expedition steps driven by VLM-generated linguistic goals. (b) Example E\&E exploration tree in Flow Lenia with \Nov{\textbf{expansions}} and \Exp{\textbf{expeditions}} steps. (c) Example expedition: E\&E minimizes the distance between the embedding of the solution's behavior (initialized to known solution \#1362) and the VLM-generated goal to uncover a new behavioral niche (solution \#2100).
    \label{fig:main}}
\end{figure*}

Discovering diverse visual patterns in continuous cellular automata (CA) represents a fundamental challenge in artificial life research \citep{chan2018lenia, reinke2019intrinsically, mordvintsev2020growing, Faldor2024TowardAO}. Despite their capacity for rich, emergent behaviors, these systems map parameters to outcomes in highly non-linear and redundant ways that defy efficient exploration. Traditional methods like Novelty Search (NS) initially uncover novel solutions but eventually stagnate when local novelty is exhausted, while random parameter search fails to penetrate the narrow regions where novelty resides. Flow Lenia, a continuous CA known for its life-like dynamics, exemplifies this challenge\,---\,its vast potential for generating intricate, self-organizing patterns remains largely untapped by conventional search techniques \citep{Plantec2022FlowLeniaTO}. Efficiently exploring these high-dimensional spaces requires strategies that can navigate through vast redundancies to identify the rare  configurations leading to emergent complexity.

To address these limitations, we propose Expedition \& Expansion (E\&E), a hybrid exploration strategy inspired by the cyclical nature of human discovery. Historically, breakthroughs emerged from alternating expeditions to uncharted territories and local expansions around new landmarks \citep{allen1972analysis,barrangou2016applications}. We translate this intuition into a computational framework: \textit{expeditions} are goal-directed search phases oriented towards distant linguistic goals hypothesizing the existence of novel behavioral patterns, while \textit{expansions} are undirected novelty-driven phases that thoroughly explore around these discoveries (Figure~\ref{fig:main}). This cyclical process mirrors human discovery: targeted expeditions probe uncharted territories, while local expansions consolidate the findings, progressively mapping out the behavioral landscape.

Existing exploration methods fall into two categories: undirected and goal-directed exploration. Undirected methods like Novelty Search (NS) \citep{lehman2011abandoning} generate diversity by prioritizing behaviors that differ from those previously encountered, gradually expanding the known territory. However, local mutations lead to a gradual exploration which may plateau when local novelty runs out or fails to bridge distant regions of the search space. Conversely, goal-directed methods such as autotelic exploration \citep{forestier2022intrinsically, colas2022autotelic} actively set behavioral targets and optimize towards them, enabling jumps to distant regions but often requiring predefined goal generators and substantial computational resources. E\&E combines both strategies, leveraging autotelic search during expeditions to target distant behaviors and using undirected novelty search to thoroughly explore their surroundings. 

The choice of representation space is critical in exploration: it determines what is perceived as novel and what goals can be effectively pursued. This framing directly influences the kinds of discoveries that are possible, constraining the outcomes of any search process. Ideally, representations should align with human perception. Traditional methods rely on hand-crafted descriptors \citep{lehman2011abandoning, mouret2015illuminating, papadopoulos2024looking}, which confine exploration to predefined features. Learned representations broaden the search space but often lack semantic alignment with human intuitions, making the novelty they capture less interpretable \citep{pere2018unsupervised, cully2019autonomous}. Interactive evolutionary computation can introduce human alignment \citep{takagi2002interactive}, however this is bottlenecked by the availability of human resources.

Foundation models can help address these limitations. Multimodal models like CLIP \citep{radford2021learning} project both images and text into a shared semantic space, enabling novelty to be computed in conceptually meaningful ways \citep{kumar2024automating, tam2022semantic}. Similarly, large language models (LLMs) have been used to generate diverse, semantically rich goals to drive autotelic exploration in reinforcement learning \citep{colas2022autotelic, colas2023augmenting, du2023guiding, zhang2023omni, pourcel2024aces}. E\&E builds on these advances by measuring novelty in semantic spaces and leveraging a Vision-Language Model (VLM) to generate linguistic goals that guide its exploration. These goals act as hypotheses about the existence of interesting behaviors in unexplored regions of the search space\,---\,much like a scientist formulates theories to orient their research. 
    
Concretely, E\&E operates through two alternating phases: expansion and expedition (Figure~\ref{fig:main}).
In the expansion phase, the algorithm applies Novelty Search (NS) to explore locally, mutating solutions that differ from previously discovered behaviors. Every $K$ iterations, E\&E initiates an expedition phase, where a VLM-generated linguistic goal directs search toward distant, semantically novel regions. By alternating between undirected local exploration and goal-directed expeditions, E\&E systematically maps out uncharted regions of the behavioral landscape, pushing beyond the limitations of purely local search methods.

We evaluate E\&E in the context of Flow Lenia, a continuous cellular automaton. Our experiments demonstrate that E\&E consistently discovers more diverse solutions compared to three baselines: Random Parameters, Random Genetic Algorithm (GA), and Novelty Search (NS). Importantly, diversity gains achieved by E\&E transfer robustly to independent embedding spaces, indicating that it does not merely overfit to CLIP’s latent structure. Furthermore, a genealogical analysis reveals that solutions originating from expeditions produce disproportionately many descendants, suggesting that they unlock new behavioral niches that drive long-term exploration.

\noindent
Our contributions are threefold: 
\begin{itemize}[left=0.5em, itemsep=1pt, topsep=3pt]
    \item We introduce Expedition and Expansion (E\&E), a novel exploration strategy that systematically combines goal-directed expeditions and undirected expansions to overcome the limitations of existing methods;
    \item We demonstrate that leveraging semantic representation spaces for both novelty computation and goal generation enables richer and more robust exploration, validated across independent embedding spaces; 
    \item We provide empirical evidence that expedition-discovered solutions are disproportionately influential in long-term exploration, serving as stepping stones for further discovery. 
\end{itemize}

Together, these findings suggest that E\&E not only expands the frontier of exploration but also reshapes the search process to uncover more diverse and meaningful behaviors.

\section{Related Work}
\label{sec:related_work}

\textbf{Exploration in cellular automata.} Exploration in cellular automata (CA) has been approached with both undirected and goal-directed methods \citep{mordvintsev2020growing, niklasson2021self-organising, hamon2024discovering}. Previous work leveraged Novelty Search and goal-directed algorithms combined with mutation-based optimization to explore Lenia using hardcoded representations \citep{reinke2019intrinsically}. Others proposed learning the space with VAEs and applying either Novelty Search or Map-Elites \citep{etcheverry2020hierarchically, Faldor2024TowardAO}. More recently, concurrent work leveraged CLIP embeddings to optimize CA behaviors towards predefined goals or novelty objectives \citep{kumar2024automating}. In contrast, E\&E dynamically generates its own goals in semantic spaces and alternates between undirected local search and targeted expeditions to escape local optima and explore new behavioral regions.

\noindent
\textbf{Representation spaces for exploration.} The effectiveness of any exploration methods hinges on representation choice. Early methods relied on hand-crafted behavioral descriptors \citep{lehman2011abandoning, mouret2015illuminating, schaul2015universal}, limiting exploration to manually defined features. Learned representations through variational autoencoders \citep{pere2018unsupervised, cully2019autonomous, etcheverry2020hierarchically, Faldor2024TowardAO} removed this constraint but often captured features misaligned with human perception. Recent advances leverage semantic representation learning with models like CLIP \citep{radford2021learning}, aligning behaviors with linguistic concepts \citep{tam2022semantic,kumar2024automating} to discover \textit{semantically} novel solutions.

\noindent
\textbf{Goal generation with large language models.} While autotelic approaches benefit from explicit goal-setting, the quality of exploration depends critically on how these goals are generated. Large language models (LLMs) have recently emerged as powerful goal generators for autotelic exploration \citep{colas2023augmenting,du2023guiding,zhang2023omni,pourcel2024aces}, leveraging their extensive linguistic and cultural knowledge to sample novel, interesting objectives. These models serve as implicit ``models of interestingness'' \citep{zhang2023omni}, drawing from the rich semantic structure of language to generate goals that align with human-interpretable concepts. Unlike earlier approaches with manually crafted or randomly generated goals, LLM-generated objectives guide exploration toward semantically meaningful regions that might remain undiscovered through undirected methods.

\noindent
\textbf{Our contribution.} E\&E the first exploration algorithm to leverage a Vision-Language Model (VLM) for dynamically generating semantic goals in Continuous Cellular Automata (CA). Unlike previous autotelic methods pursuing goals sampled from hard-coded \citep{reinke2019intrinsically}, or learned but purely visual representations \citep{etcheverry2020hierarchically, Faldor2024TowardAO}, E\&E uses VLM-generated hypotheses to guide search toward semantically meaningful behaviors. These goals are expressed as linguistic descriptions that encapsulate high-level behavioral targets, enabling direct optimization in a way that is aligned with human intuition. By combining this goal-directed exploration with Novelty Search in a shared semantic space, E\&E efficiently escapes local optima, systematically expanding the search frontier and uncovering a richer diversity of behaviors. 

\section{Methods}
\label{sec:methods}

We introduce Expedition \& Expansion (E\&E), a hybrid exploration strategy designed to efficiently navigate the high-dimensional behavioral space of Flow Lenia. E\&E operates in two alternating phases: an \textit{expansion phase} driven by undirected Novelty Search in a semantic representation space, and an \textit{expedition phase} guided by VLM-generated semantic goals. The visualization and code for this project is included in the URL on the first page.


\subsection{Flow Lenia: from parameters to behaviors}
\label{sec:flowlenia}
Flow Lenia is a continuous cellular automaton that extends the original Lenia framework with mass conservation and dynamic parameter localization, enabling the emergence of complex, life-like behaviors \citep{Plantec2022FlowLeniaTO}. Each instance is parameterized by a set of kernel parameters (number of concentric rings, their sizes, heights, center positions, and widths) and growth function parameters (mean and standard deviation of the growth curve), see Table 1 of \cite{Plantec2022FlowLeniaTO}. Together, these parameters govern local interactions that shape the state of the automaton over time. These 235 parameters, denoted as $\theta$, are \textit{sampled} during archive initialization, \textcolor{nov}{\textit{mutated}} during \textcolor{nov}{expansion}, and \textcolor{exp}{\textit{optimized}} during \textcolor{exp}{expeditions}. For a given set of parameters $\theta$, we simulate Flow Lenia for $T=500$ steps and define the resulting final state\,---\,a 128$\times$128 image\,---\,as its \textit{behavior}.


\subsection{Representing behaviors with semantic embeddings}
\label{sec:clip}
Directly comparing Flow Lenia behaviors (raw 128$\times$128 images) using pixel-wise distance metrics would be meaningless: small pixel differences do not reflect meaningful behavioral changes, and visually similar patterns may be arbitrarily distant in pixel space. To obtain more meaningful novelty measures, we leverage multimodal embeddings from CLIP, a vision-language model trained on large-scale image-text pairs \citep{radford2021learning}. CLIP maps both images and text into a shared latent space that captures high-level semantic features, allowing conceptually similar behaviors to cluster together. We define the embedding function as $\phi$ such that $\text{embedding}=\phi(\text{behavior})$. The resulting latent vector serves as a compact, semantically meaningful representation of the behavior often called \textit{behavioral descriptor} in evolutionary research \citep{cully2017quality}, and \textit{outcome} in reinforcement learning \citep{colas2022autotelic}. During both the expansions and expeditions, novelty and goal alignment are computed as cosine distances in this structured space, allowing for efficient and semantically grounded exploration \citep{tam2022semantic,kumar2024automating}.


\subsection{Expedition \& Expansion}
\label{sec:e_and_e}

\textbf{Archive initialization.} The exploration process begins with a seeding phase, where we sample 1,000 random parameter configurations $\theta$ from Flow Lenia's search space and simulate them to obtain corresponding behaviors. Once embedded, we use the resulting 1,000 solutions $(\theta, \text{behavior}, \text{embedding})$ to populate an archive $A$, serving as the initial stepping stones for future expansions and expeditions.

\noindent
\textbf{\textcolor{nov}{Expansion phase with Novelty Search}.} The expansion phase aims to explore local regions of the parameter space by identifying and mutating novel behaviors. As is standard, we define novelty as the average cosine distance of a solution's embedding to its $k\,=\,10$ nearest neighbors \citep{lehman2011abandoning, etcheverry2020hierarchically, Faldor2024TowardAO}: 
\begin{equation*}
    \textsc{NOV}(\theta)=\frac{1}{k} \sum_{e'\in \mathcal{N}^k(e)} (1 - sim(e,\,e')),
\end{equation*}
where $e$ is the embedding of parameters $\theta$, $\mathcal{N}^k(e)$ is the set of its $k$ nearest neighbors in embedding space, and $cos\_sim$ is the cosine similarity. At each iteration, Novelty Search samples a solution from the archive in proportion to its novelty with bias $\alpha$ ($p\propto \textsc{NOV}^\alpha$) and mutates its parameters $\theta_\text{novel}$ with additive Gaussian noise to obtain a new set of parameters: $\theta'=\textsc{Mutate}(\theta_\text{novel})$. After simulating this set of parameters in Flow Lenia, the new behavior is embedded, evaluated for novelty, and the complete solution ($\theta, \text{behavior}, \text{embedding}$) is added to the archive. This iterative process steadily pushes the frontier of local exploration, setting the stage for the expedition phase to uncover more distant regions.

\noindent
\textbf{\textcolor{exp}{Expedition phase with goal-directed search.}} The expedition phase aims to overcome local novelty plateaus by directing exploration towards distant, under-explored regions of the behavioral space. This is achieved through autotelic search: goal generation and goal-directed optimization \citep{colas2022autotelic}. Every $K$ steps, the algorithm shifts from local exploration to a targeted search by prompting a Vision-Language Model (VLM) to generate a linguistic goal to pursue (Open-AI's \texttt{o4-mini}).

To ground goal generation, we provide as context the list of past generated goals as well as a sample of $n=25$ behaviors (images) found in the archive, in proportion to their novelty with bias $\alpha$. We seed the list with six hand-crafted goals that provide the VLM with concrete examples before it begins generating its own objectives from context. We then instruct the VLM to analyze the characteristics that are either \textbf{shared} or \textbf{unique} across the sampled images, and to consider \textit{adjacent possibles}\,---\,a form of chain-of-thought prompting \citep{wei2022chain}. Finally, we instruct the VLM to generate a 5-15 words goal description to direct the expedition (see examples in Section~\nameref{sec:results}). This prompting technique encourages the VLM to hypothesize about unseen but plausible behaviors, mirroring the process of scientific hypothesis generation.

The linguistic goal is then embedded into the same CLIP latent space as Flow Lenia behaviors: $e_\text{goal}=\phi(\text{goal})$. To optimize towards this goal, we locate the nearest neighbor to $e_\text{goal}$ in the archive and use its parameters $\theta_\text{NN}$ to initialize a goal-directed optimization process. We use sep-CMA-ES, an evolutionary optimization algorithm that we run for 350 optimization steps with a population of $N=16$, $\sigma_{\text{init}}=0.1$~\citep{ros2008simple}. We formalize the objective of this goal-directed process as the minimization of the cosine distance between the behavior's embedding and the goal embedding:
\begin{equation*}
    \theta'= \argmin_\theta (1 - sim (\phi(\text{behavior}(\theta)), e_\text{goal})).
\end{equation*}
After optimization, the new solution is added to the archive, contributing novel candidate stepping stones for further expansions and expeditions. 

\noindent
\textbf{Closing the loop.} After an initial seeding phase of $S$ iterations, E\&E runs for $N$ iterations, primarily performing expansion steps. Every $K$ iterations, it switches to an expedition step, introducing goal-directed exploration before resuming expansion. The complete pseudo-code is presented in Algorithm~\ref{alg:dens}.

\input{alg}

\subsection{Baselines and diversity measure}
\label{sec:baselines}

We compare Expedition and Expansion (E\&E) against three baselines:
\begin{itemize}[left=0.5em, itemsep=1pt, topsep=3pt]
    \item \textbf{Random Parameter Search}: Parameters are sampled uniformly from the parameter space, mirroring the initialization of the archive.
    \item \textbf{Random Genetic Algorithm (GA)}: Parameters are sampled randomly from the archive and mutated with Gaussian noise.
    \item \textbf{Novelty Search (NS)}: This ablates the expedition phase to focus on expansion only, implemented by the Novelty Search algorithms using different novelty biases $\alpha$. 
\end{itemize}

\noindent
We aim to generate diversity that is meaningful from a human perspective. To evaluate the diversity of an archive of solutions, we compute the average pairwise distance between embeddings in the archive. This is done separately in two distinct representation spaces: CLIP and DINO \citep{radford2021learning,caron2021emerging}. CLIP is a multimodal model trained on large-scale image-text pairs that capture semantically structured representations \citep{radford2021learning}. This alignment with linguistic descriptions provides a human-centric perspective, reflecting visual features that people tend to describe. However, CLIP's alignment is not perfect; optimization processes can exploit specific latent features to artificially inflate novelty scores for behaviors that may not seem novel to human observers. To guard against this, we additionally measure diversity in DINO, a self-supervised model trained independently of linguistic supervision \citep{caron2021emerging}. This complementary evaluation helps ensure that the discovered novelty is not just an artifact of CLIP's biases but represents genuinely diverse behaviors that generalize across different representation spaces.

\section{Results}
\label{sec:results}
We evaluate the effectiveness of the Expedition \& Expansion (E\&E) strategy in exploring the high-dimensional behavioral space of Flow Lenia. Our analysis focuses on three main aspects: (1) \textbf{diversity analysis}, evaluating the impact of novelty bias calibration, expedition frequency, and baseline comparisons; (2) \textbf{expedition analysis}, examining goals generated during expeditions and how expeditions optimize towards these goals; and (3) \textbf{genealogical analysis}, tracing the long-term influence of expedition-discovered solutions and their role in unlocking new behavioral clusters. Our results indicate that E\&E consistently uncovers richer and more diverse solutions, pushing exploration beyond the limitations of local Novelty Search. An interactive visualization of the results can be found at \href{https://incognito-researcher.github.io/visu_flow_direct/}{here}.

\begin{figure*}[!h]
    \centering
    \includegraphics[width=\linewidth]{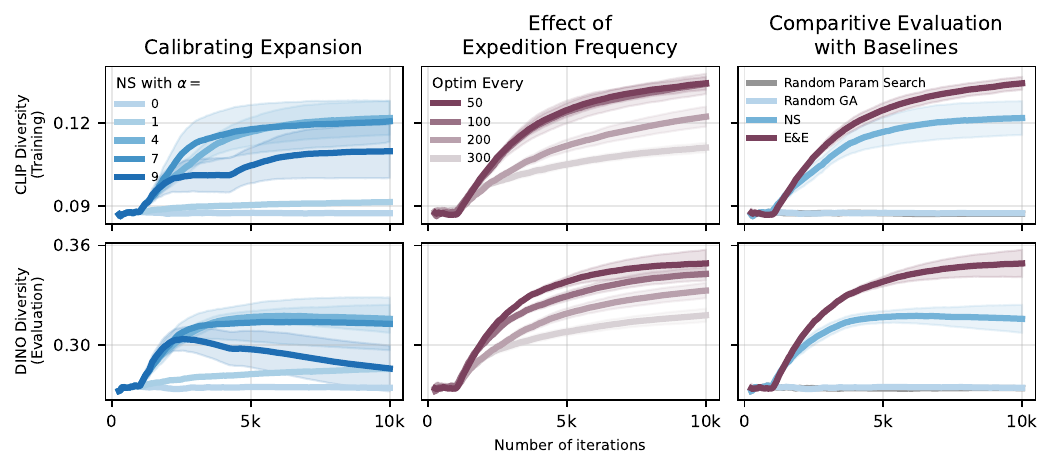}
    \caption{
    \textbf{Diversity analysis.} Lines show the diversity of discovered solutions as measured by the average pairwise distance of CLIP (top) or DINO (bottom) solution embeddings (N=5 seeds, mean $\pm$ standard error). 
    \textit{Left:} Diversity discovered by Novelty Search (NS) under different novelty biases ($p=\textsc{nov}^\alpha$).
    \textit{Center:} Diversity discovered by E\&E's under different expedition frequencies ($1/K$).
    \textit{Right:} Diversity discovered by E\&E vs baseline methods. 
    }
    \label{fig:diversity}
\end{figure*}

\subsection{Diversity analysis}
\label{sec:res_diversity}

\textbf{Calibrating the expansion phase.} The expansion phase in E\&E relies on Novelty Search (NS) to explore locally by mutating high-novelty solutions. A key parameter in this process is the novelty bias coefficient $\alpha$, which controls how strongly the search prioritizes novel behaviors: $p\sim\textsc{nov}^\alpha$. To identify the optimal setting, we perform a parameter sweep of $\alpha$ from 0 (random GA) to 9, measuring diversity as the average pairwise distance of embeddings in either CLIP or DINO spaces (Figure~\ref{fig:diversity}, first column).

The default setting of $\alpha=1$ shows minimal improvement over the random genetic algorithm ($\alpha=0$), likely due to the limited variance in cosine distances measured in these high-dimensional embeddings. Increasing $\alpha$ to 4 amplifies novelty signals, effectively driving exploration towards more diverse behaviors. However, setting $\alpha > 4$ introduces instability: exploration inflates in CLIP's latent space but collapses in DINO, indicating that NS may be overfitting to the particular features of CLIP. We select $\alpha=4$ as it optimally balances local novelty discovery and robustness across representation spaces, ensuring more stable and meaningful expansion.

\noindent
\textbf{Effect of expedition frequency in E\&E.} A central feature of E\&E is its periodic shift from local Novelty Search to goal-directed exploration during expedition phases. The frequency of these expeditions is controlled by parameter $K$, which determines the number of expansion iterations between each semantic search. We evaluate the impact of $K$ on exploration diversity by testing values of $K$ in \{50, 100, 200, 300\} and measuring the resulting diversity discovered in both CLIP and DINO spaces (Figure~\ref{fig:diversity}, second column).

Our results show that the frequency of expeditions ($1/K$) significantly affects exploration diversity. Higher values of $K$ (200, 300), corresponding to fewer expeditions, seem to perform worse than pure Novelty Search (NS) in CLIP space, while performing on par with NS in DINO space. This suggests that sparse expeditions do not meaningfully enhance diversity, but may help reduce NS's overfitting to CLIP features. In contrast, lower values of $K$ (50, 100) consistently improve diversity, especially in DINO space, indicating that frequent expeditions introduce more general diversity rather than overfitting to CLIP-specific features. However, diminishing returns appear when comparing $K=100$ and $K=50$, suggesting that increasing expedition frequency beyond a certain point does not significantly enhance exploration. We select $K=50$ as it strikes an optimal balance between diversity gains and computational efficiency.

\noindent
\textbf{Comparative evaluation with baselines.} We compare E\&E against three standard baselines: Random Parameter Search, Random Genetic Algorithm (GA), and Novelty Search (NS) with the optimal bias coefficient $\alpha=4$. Random Parameter Search samples parameters uniformly from the search space, while Random GA selects random solutions from the archive and mutates them with Gaussian noise before adding the results back to the archive. NS, in turn, focuses purely on undirected novelty-driven exploration. Note than Random GA is equivalent to running NS with $\alpha=0$, i.e., without a novelty bias.

Figure~\ref{fig:diversity} (third column) shows the diversity of solutions discovered by each method across 10,000 iterations in the CLIP and DINO spaces. E\&E consistently outperforms all baselines, uncovering more diverse behaviors and maintaining exploration over time. This improvement is particularly evident in DINO space, indicating that E\&E does not merely exploit CLIP-specific features but genuinely uncovers new behavioral niches. 

Unlike NS, which tends to plateau after about 3,000 iterations, E\&E's semantic expeditions allow it to continuously extend the search frontier. As supported by the genealogical study presented in Section~\nameref{sec:geneology}, the periodic introduction of goal-directed exploration injects diversity into regions that purely undirected methods fail to reach. This highlights E\&E's ability to break through local novelty barriers, unlocking new behavioral niches that drive longer-term exploration.


\subsection{Expedition analysis}
\label{sec:res_expedition}
During expedition phases, E\&E leverages VLMs to generate linguistic goals that direct exploration toward semantically novel behaviors. Table~\ref{tab:goals} presents examples of linguistic goals generated during expeditions. The goals tend to fall into two main themes: (i) geometric primitives (e.g., \textit{a green hexagonal grid with a red and blue spiral at the center}), or (ii) natural forms such as flowers, shells (e.g., \textit{a photo of a spiky sea urchin with green spines radiating from a red spherical body}). These goals tend to get longer and more compositional as the archive grows. Simple geometric goals are more easily reached through optimization and tend to spawn larger descendant clusters, whereas naturalistic goals are harder to reach through optimization. 

\rowcolors{2}{white}{gray!25}
\renewcommand{\arraystretch}{1.2}
\begin{table}[H]
    \centering
    \small
    \begin{tabular}{cp{0.75\linewidth}}
    \toprule
        \textbf{Iteration} & \textbf{Linguistic Goal} \\ 
        \midrule
        \rowcolor{seed_light}
        1100 & a pink square \hfill\textit{(pre-defined)}\\ 
        1650 & a photo of a jellyfish with trailing tentacles\\
        1850 & a photo of a butterfly with red wings and green dot patterns\\
        2150 & a photo of a four-leaf clover with bright green heart-shaped leaves\\
        3550 & a photo of a three-armed spiral symbol with red and green arcs\\
        6350 & a photo of a yin-yang symbol with red and blue halves and green dots\\
        9750 & a photo of a lava lamp with red blobs and green droplets in blue fluid\\
        \bottomrule
    \end{tabular}
    \caption{\textbf{Examples of linguistic goals generated during expeditions (one seed).} }
    \label{tab:goals}
\end{table}

\begin{figure}[h]
    \centering
    \includegraphics{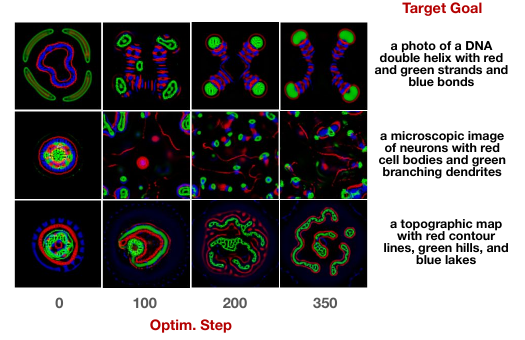}
    \caption{
    \textbf{Goal-directed expeditions.} 
    Starting from the nearest neighbor in the archive, sep-CMA-ES optimizes behaviors towards VLM-generated linguistic goals. Shown are key steps (0, 100, 200, 350) illustrating gradual alignment and novel pattern discovery.}
    \label{fig:optim}
\end{figure}

For each linguistic goal, E\&E initializes a goal-directed optimization process using sep-CMA-ES, seeded with the nearest neighbor solution found in the archive \citep{ros2008simple}. Figure~\ref{fig:optim} illustrates the optimization trajectory for three different expedition phases, showing the gradual alignment of behaviors towards the specified goals. While not all optimization traces achieve perfect alignment with their linguistic descriptions, this is not necessarily a failure. The primary purpose of expeditions is to break out of local neighborhoods and push the search into novel regions of the behavioral space. The next section shows that even partial goal achievement often produces stepping stones leading to further exploration during the subsequent expansion phases. 

\subsection{Genealogical analysis}
\label{sec:geneology}
To evaluate the long-term impact of expedition-discovered solutions, we perform a genealogical analysis that traces the descendants of solutions introduced by expedition steps. Despite representing only a minority of the total search iterations, solutions originating from expeditions exhibit a disproportionate influence on the final archive. For $K=50$, E\&E executes only 179 expeditions over 10,000 iterations. Under a purely random genetic algorithm (GA) model, we would expect the progeny of these 179 solutions to represent $4.64\% \pm 0.36\%$ of the final archive, assuming uniform selection pressure. In contrast, empirical results across five independent seeds reveal that expedition-discovered solutions constitute $42.2\% \pm 13.7\%$ of the final archive\,---\,an order of magnitude greater than the random expectation. 

To contextualize this effect, we compare the average progeny of expedition-discovered solutions with those of parent solutions generated within a $\pm10$ iteration window of each expedition. These contemporaneous solutions average only $3.45\% \pm 1.6\%$ of descendants in the final archive, further highlighting the uniquely generative role of expeditions. This overrepresentation suggests that expedition-discovered solutions systematically unlock new regions of the behavioral space, serving as robust stepping stones for subsequent novelty-driven exploration.

\begin{figure}[t!]
    \centering
    \includegraphics[width=\linewidth]{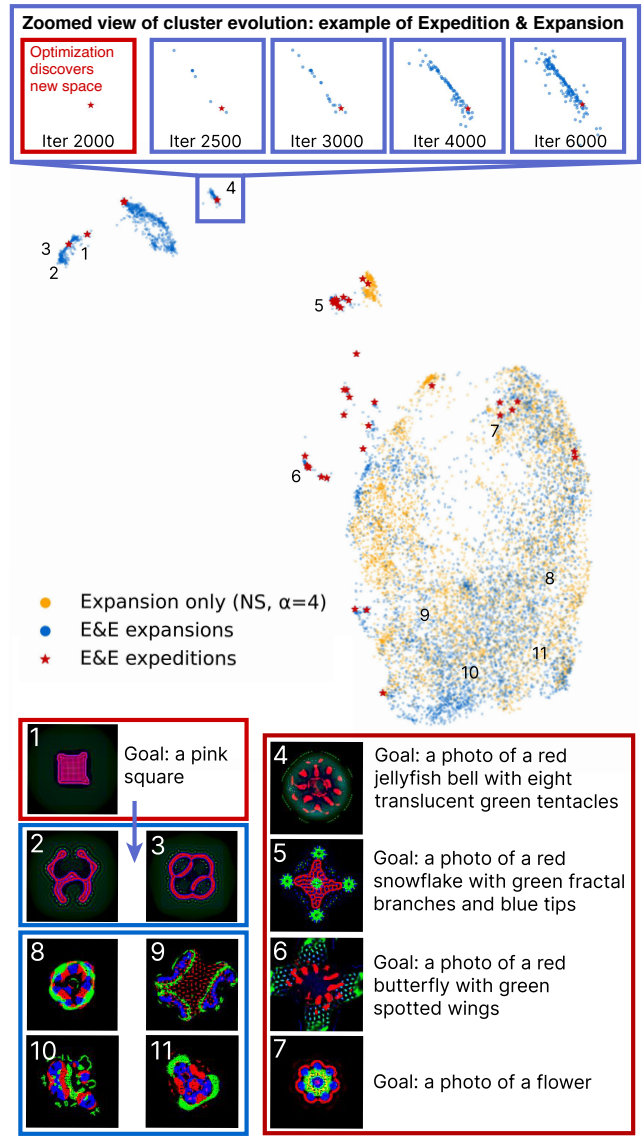}
    \caption{
    \textbf{Expeditions pioneers new novelty niches.} We embed the final archives in 2D with UMAP. Orange points come from a expansion-only (NS) run; blue points from E\&E, with red stars marking behaviors created during expedition phases. Top: example of an expedition seeding a new cluster beyond previously explored regions. Bottom: Examples behaviors discovered by expansions (blue) and expeditions (red).%
    }
    \label{fig:umap}
\end{figure}

Figure~\ref{fig:umap} presents a UMAP projection of the behavioral space explored by E\&E and the NS baseline. Each point represents a discovered behavior, with color coding for expansion-only (NS) (in blue) and E\&E (in red). Stars indicate solutions obtained on expedition steps. The visualization reveals that expeditions consistently discover new frontiers in the search space, populating regions that are largely unexplored by NS alone. These newly unlocked areas are not isolated: they serve as stepping stones that Novelty Search leverages to further explore local variations, leading to a denser population of behaviors in previously unreachable niches. This demonstrates the key advantage of E\&E's hybrid strategy: goal-directed expeditions push beyond local attractors, creating fertile ground for NS to expand upon.

\section{Discussion}

Expedition \& Expansion introduces a novel exploration paradigm that systematically discovers truly different and semantically rich patterns in continuous cellular automata. By alternating between undirected novelty search and directed expeditions guided by linguistic hypotheses, E\&E uncovers diverse behaviors that are not just novel in representation space but meaningfully distinct, as evidenced by both diversity metrics and qualitative examples. This structured alternation pushes beyond the limitations of purely stochastic exploration, enabling the discovery of new behavioral niches that drive long-term exploration.

\textbf{Limitations.} While E\&E significantly expands the exploration frontier in continuous cellular automata, three main limitations remain. First, its dual-phase mechanism introduces substantial computational costs. Expedition steps require large VLMs (\texttt{o4-mini}) for goal generation and sep-CMA-ES for optimization, averaging 3 minutes per step on an A100 GPU, against about 1 second for expansion steps. This cost accumulates with frequent expeditions, limiting scalability to larger parameter spaces. This could be mitigated through adaptive scheduling or early stopping criteria based on convergence signals during optimization. 

Second, E\&E is susceptible to novelty hacking: optimization sometimes inflates novelty scores that do not translate into meaningful differences in other representation spaces. Mitigating this effect may require ensembling multiple representation spaces or discounting novelty for direct descendants to reduce overexploitation. 

Third, the VLM-based goal generation lacks diversity and temporal adaptation. Despite linguistic variability, goals remain largely static across iterations, behaving more like a fixed generator than an adaptive explorer. Ideally, the VLM should react to past search outcomes, progressively shifting its sampling distribution toward different niches in light of past successes and failures. Integrating feedback loops that evaluate the distribution of generated goals over time, along with meta-intrinsic rewards, could enable adaptive, curriculum-like exploration aimed at maximizing learning progress in goal-achievement \citep{kaplan_maximizing_2004} or goal diversity \citep{pong2019skew}. This could be achieved by feeding back information on the success rate and novelty of previously attempted goals to promote a form of self-improving curriculum learning.

\textbf{Future work.} Building on E\&E's exploration paradigm, several directions emerge for extending its capabilities. First, integrating temporal embeddings could enable E\&E to capture not just static behavioral snapshots, but also the dynamics of behaviors as they evolve over time. Current representation spaces (CLIP, DINO) are limited to single-frame embeddings, missing critical temporal information that may reveal long-term stability or phase transitions in behaviors. Video-based models or self-supervised spatiotemporal embeddings could address this gap, allowing E\&E to discover temporally extended phenomena and richer dynamical patterns \citep{tong2022videomae, askvideos2024videoclip}.

Second, introducing interactive exploration represents an exciting frontier. E\&E’s goal generation mechanism, currently driven autonomously by VLM prompts, could be augmented with real-time human interaction. Researchers or designers could inject high-level goals or preferences during exploration, steering expeditions toward regions of interest or away from known failure modes \citep{etcheverry2020hierarchically}. This collaborative form of exploration could accelerate the discovery of novel patterns and allow for guided, hypothesis-driven search in real time. Alternatively, meta-learning strategies can be employed onto the goal-generating process itself in order to optimize for goal trajectories/curricula that maximize learning progress \citep{lopes2012exploration}.

Finally, extending E\&E to other domains could unlock its structured exploration paradigm in diverse settings. Applications like generative design \citep{clune2011evolving}, crystal structure prediction \citep{janmohamed2024multi}, robot skill learning \citep{colas2020scaling}, or procedural content generation \citep{gravina2019procedural}, where the behavioral space can be searched by mutation-based or goal-directed exploration methods, could all benefit from the dual mechanisms of semantic-driven expeditions and novelty-based expansions.

\section{Acknowledgements}
Cédric Colas acknowledges funding from the European Union’s Horizon 2020 research and innovation programme under the Marie Skłodowska-Curie grant agreement No 101065949.

\footnotesize
\bibliographystyle{apalike}

\end{document}

%% file: alg.tex
\begingroup
\removelatexerror 
\begin{tcolorbox}[colback=gray!05, boxrule=1pt,
                  left=3pt,right=0pt,top=2pt,bottom=2pt,
                  width=\columnwidth]   
\begin{algorithm}[H]                    
\caption{Expedition \& Expansion algorithm} \label{alg:dens}

\footnotesize
\SetAlgoLined
\DontPrintSemicolon
\KwIn{$N$ iters, $S$ init solutions, expedition every $K$ iter,\;$Sim:\theta\to \text{behavior}$ --- simulation engine,\;$\phi: \text{behavior} \to \text{embedding}$ --- behavioral descr. (CLIP)}
\KwOut{\;$A$: archive of (param, embedding) pairs $(\theta,\,\phi(Sim(\theta)))$\; $G$: goal archive}
$A \gets \emptyset$                                         \tcp*{init archive}
$G \gets$ 6 pre-defined goals                                 \tcp*[f]{init goals}

                                            
\For{$i\gets1$ \KwTo $N$}{%
\uIf(\tcp*[f]{seeding}){$i \leq S$}{
$\theta \gets \textsc{Rand}()$\;     
}
\Nov{                             
  \ElseIf(\tcp*[f]{expansion (NS)}){$i \bmod K \neq 0$}{
    $\theta_{\text{novel}} \gets$ \textsc{HighNovelty}(A,\,n=1)\;
    $\theta \gets \textsc{Mutate}(\theta_{\text{novel}})$\;   
        \def\algorithmicendif{} 
        }
  }
\Exp{
  \Else(\tcp*[f]{expedition}){%
     $\text{goal} \gets \textsc{VLM}(\textsc{HighNovelty}(A,\,10),\,G)$\;   
     $G \gets G \cup \text{goal}$\;
     $\theta_{\text{NN}} \gets \textsc{NearestNeighbor}(A, \text{goal})$\;
     $\theta \gets \textsc{sep-CMAES}(\phi, \theta_{\text{NN}}, \text{goal})$\;
     }
 }
  $\text{behavior} \gets Sim(\theta)$  \tcp*[f]{run simulation}\;
  $\text{embedding} \gets \phi(\text{behavior})$  \tcp*[f]{embed behavior}\;
  $A \gets A \cup (\theta,\,\text{embedding})$ \tcp*[f]{add to archive}
}
  


\end{algorithm}
\vspace{-0.5em}
\end{tcolorbox}
\endgroup

%% file: main.bbl
\begin{thebibliography}{}

\bibitem[Allen, 1972]{allen1972analysis}
Allen, J.~L. (1972).
\newblock An analysis of the exploratory process: The lewis and clark expedition of 1804-1806.
\newblock {\em Geographical Review}, pages 13--39.

\bibitem[AskVideos, 2024]{askvideos2024videoclip}
AskVideos (2024).
\newblock Askvideos-videoclip: Language-grounded video embeddings.
\newblock GitHub.

\bibitem[Barrangou and Doudna, 2016]{barrangou2016applications}
Barrangou, R. and Doudna, J.~A. (2016).
\newblock Applications of crispr technologies in research and beyond.
\newblock {\em Nature biotechnology}, 34(9):933--941.

\bibitem[Caron et~al., 2021]{caron2021emerging}
Caron, M., Touvron, H., Misra, I., J{\'e}gou, H., Mairal, J., Bojanowski, P., and Joulin, A. (2021).
\newblock Emerging properties in self-supervised vision transformers.
\newblock In {\em Proceedings of the IEEE/CVF international conference on computer vision}, pages 9650--9660.

\bibitem[Chan, 2018]{chan2018lenia}
Chan, B. W.-C. (2018).
\newblock Lenia-biology of artificial life.
\newblock {\em arXiv preprint arXiv:1812.05433}.

\bibitem[Clune and Lipson, 2011]{clune2011evolving}
Clune, J. and Lipson, H. (2011).
\newblock Evolving 3d objects with a generative encoding inspired by developmental biology.
\newblock {\em ACM SIGEVOlution}, 5(4):2--12.

\bibitem[Colas et~al., 2022]{colas2022autotelic}
Colas, C., Karch, T., Sigaud, O., and Oudeyer, P.-Y. (2022).
\newblock Autotelic agents with intrinsically motivated goal-conditioned reinforcement learning: a short survey.
\newblock {\em Journal of Artificial Intelligence Research}, 74:1159--1199.

\bibitem[Colas et~al., 2020]{colas2020scaling}
Colas, C., Madhavan, V., Huizinga, J., and Clune, J. (2020).
\newblock Scaling map-elites to deep neuroevolution.
\newblock In {\em Proceedings of the 2020 Genetic and Evolutionary Computation Conference}, pages 67--75.

\bibitem[Colas et~al., 2023]{colas2023augmenting}
Colas, C., Teodorescu, L., Oudeyer, P.-Y., Yuan, X., and C{\^o}t{\'e}, M.-A. (2023).
\newblock Augmenting autotelic agents with large language models.
\newblock In {\em Conference on Lifelong Learning Agents}, pages 205--226. PMLR.

\bibitem[Cully, 2019]{cully2019autonomous}
Cully, A. (2019).
\newblock Autonomous skill discovery with quality-diversity and unsupervised descriptors.
\newblock In {\em Proceedings of the Genetic and Evolutionary Computation Conference}, pages 81--89.

\bibitem[Cully and Demiris, 2017]{cully2017quality}
Cully, A. and Demiris, Y. (2017).
\newblock Quality and diversity optimization: A unifying modular framework.
\newblock {\em IEEE Transactions on Evolutionary Computation}, 22(2):245--259.

\bibitem[Du et~al., 2023]{du2023guiding}
Du, Y., Watkins, O., Wang, Z., Colas, C., Darrell, T., Abbeel, P., Gupta, A., and Andreas, J. (2023).
\newblock Guiding pretraining in reinforcement learning with large language models.
\newblock In {\em International Conference on Machine Learning}, pages 8657--8677. PMLR.

\bibitem[Etcheverry et~al., 2020]{etcheverry2020hierarchically}
Etcheverry, M., Moulin-Frier, C., and Oudeyer, P.-Y. (2020).
\newblock Hierarchically organized latent modules for exploratory search in morphogenetic systems.
\newblock {\em Advances in Neural Information Processing Systems}, 33:4846--4859.

\bibitem[Faldor and Cully, 2024]{Faldor2024TowardAO}
Faldor, M. and Cully, A. (2024).
\newblock Toward artificial open-ended evolution within lenia using quality-diversity.
\newblock {\em ArXiv}, abs/2406.04235.

\bibitem[Forestier et~al., 2022]{forestier2022intrinsically}
Forestier, S., Portelas, R., Mollard, Y., and Oudeyer, P.-Y. (2022).
\newblock Intrinsically motivated goal exploration processes with automatic curriculum learning.
\newblock {\em Journal of Machine Learning Research}, 23(152):1--41.

\bibitem[Gravina et~al., 2019]{gravina2019procedural}
Gravina, D., Khalifa, A., Liapis, A., Togelius, J., and Yannakakis, G.~N. (2019).
\newblock Procedural content generation through quality diversity.
\newblock In {\em 2019 IEEE Conference on Games (CoG)}, pages 1--8. IEEE.

\bibitem[Hamon et~al., 2024]{hamon2024discovering}
Hamon, G., Etcheverry, M., Chan, B. W.-C., Moulin-Frier, C., and Oudeyer, P.-Y. (2024).
\newblock Discovering sensorimotor agency in cellular automata using diversity search.
\newblock {\em arXiv preprint arXiv:2402.10236}.

\bibitem[Janmohamed et~al., 2024]{janmohamed2024multi}
Janmohamed, H., Wolinska, M., Surana, S., Pierrot, T., Walsh, A., and Cully, A. (2024).
\newblock Multi-objective quality-diversity for crystal structure prediction.
\newblock In {\em Proceedings of the Genetic and Evolutionary Computation Conference}, pages 1273--1281.

\bibitem[Kaplan and Oudeyer, 2004]{kaplan_maximizing_2004}
Kaplan, F. and Oudeyer, P.-Y. (2004).
\newblock Maximizing {Learning} {Progress}: {An} {Internal} {Reward} {System} for {Development}.
\newblock In {\em Embodied artificial intelligence}, pages 259--270. Springer.

\bibitem[Kumar et~al., 2024]{kumar2024automating}
Kumar, A., Lu, C., Kirsch, L., Tang, Y., Stanley, K.~O., Isola, P., and Ha, D. (2024).
\newblock Automating the search for artificial life with foundation models.

\bibitem[Lehman and Stanley, 2011]{lehman2011abandoning}
Lehman, J. and Stanley, K.~O. (2011).
\newblock Abandoning objectives: Evolution through the search for novelty alone.
\newblock {\em Evolutionary Computation}, 19(2):189--223.

\bibitem[Lopes et~al., 2012]{lopes2012exploration}
Lopes, M., Lang, T., Toussaint, M., and Oudeyer, P.-Y. (2012).
\newblock Exploration in model-based reinforcement learning by empirically estimating learning progress.
\newblock {\em Advances in neural information processing systems}, 25.

\bibitem[Mordvintsev et~al., 2020]{mordvintsev2020growing}
Mordvintsev, A., Randazzo, E., Niklasson, E., and Levin, M. (2020).
\newblock Growing neural cellular automata.
\newblock {\em Distill}, 5(2):e23.

\bibitem[Mouret and Clune, 2015]{mouret2015illuminating}
Mouret, J.-B. and Clune, J. (2015).
\newblock Illuminating search spaces by mapping elites.
\newblock {\em arXiv preprint arXiv:1504.04909}.

\bibitem[Niklasson et~al., 2021]{niklasson2021self-organising}
Niklasson, E., Mordvintsev, A., Randazzo, E., and Levin, M. (2021).
\newblock Self-organising textures.
\newblock {\em Distill}.
\newblock https://distill.pub/selforg/2021/textures.

\bibitem[Papadopoulos et~al., 2024]{papadopoulos2024looking}
Papadopoulos, V., Doat, G., Renard, A., and Hongler, C. (2024).
\newblock Looking for complexity at phase boundaries in continuous cellular automata.
\newblock In {\em Proceedings of the Genetic and Evolutionary Computation Conference Companion}, pages 179--182.

\bibitem[P{\'e}r{\'e} et~al., 2018]{pere2018unsupervised}
P{\'e}r{\'e}, A., Forestier, S., Sigaud, O., and Oudeyer, P.-Y. (2018).
\newblock Unsupervised learning of goal spaces for intrinsically motivated goal exploration.
\newblock {\em arXiv preprint arXiv:1803.00781}.

\bibitem[Plantec et~al., 2022]{Plantec2022FlowLeniaTO}
Plantec, E., Hamon, G., Etcheverry, M., Oudeyer, P.-Y., Moulin-Frier, C., and Chan, B. W.-C. (2022).
\newblock Flow-lenia: Towards open-ended evolution in cellular automata through mass conservation and parameter localization.
\newblock {\em The 2023 Conference on Artificial Life}.

\bibitem[Pong et~al., 2019]{pong2019skew}
Pong, V.~H., Dalal, M., Lin, S., Nair, A., Bahl, S., and Levine, S. (2019).
\newblock Skew-fit: State-covering self-supervised reinforcement learning.
\newblock {\em arXiv preprint arXiv:1903.03698}.

\bibitem[Pourcel et~al., 2024]{pourcel2024aces}
Pourcel, J., Colas, C., Molinaro, G., Oudeyer, P.-Y., and Teodorescu, L. (2024).
\newblock Aces: Generating a diversity of challenging programming puzzles with autotelic generative models.
\newblock {\em Advances in Neural Information Processing Systems}, 37:67627--67662.

\bibitem[Radford et~al., 2021]{radford2021learning}
Radford, A., Kim, J.~W., Hallacy, C., Ramesh, A., Goh, G., Agarwal, S., Sastry, G., Askell, A., Mishkin, P., Clark, J., et~al. (2021).
\newblock Learning transferable visual models from natural language supervision.
\newblock In {\em International conference on machine learning}, pages 8748--8763. PmLR.

\bibitem[Reinke et~al., 2019]{reinke2019intrinsically}
Reinke, C., Etcheverry, M., and Oudeyer, P.-Y. (2019).
\newblock Intrinsically motivated discovery of diverse patterns in self-organizing systems.
\newblock {\em arXiv preprint arXiv:1908.06663}.

\bibitem[Ros and Hansen, 2008]{ros2008simple}
Ros, R. and Hansen, N. (2008).
\newblock A simple modification in cma-es achieving linear time and space complexity.
\newblock In {\em International conference on parallel problem solving from nature}, pages 296--305. Springer.

\bibitem[Schaul et~al., 2015]{schaul2015universal}
Schaul, T., Horgan, D., Gregor, K., and Silver, D. (2015).
\newblock Universal value function approximators.
\newblock In {\em International conference on machine learning}, pages 1312--1320. PMLR.

\bibitem[Takagi, 2002]{takagi2002interactive}
Takagi, H. (2002).
\newblock Interactive evolutionary computation: Fusion of the capabilities of ec optimization and human evaluation.
\newblock {\em Proceedings of the IEEE}, 89(9):1275--1296.

\bibitem[Tam et~al., 2022]{tam2022semantic}
Tam, A., Rabinowitz, N., Lampinen, A., Roy, N.~A., Chan, S., Strouse, D., Wang, J., Banino, A., and Hill, F. (2022).
\newblock Semantic exploration from language abstractions and pretrained representations.
\newblock {\em Advances in neural information processing systems}, 35:25377--25389.

\bibitem[Tong et~al., 2022]{tong2022videomae}
Tong, Z., Song, Y., Wang, J., and Wang, L. (2022).
\newblock Videomae: Masked autoencoders are data-efficient learners for self-supervised video pre-training.
\newblock {\em Advances in neural information processing systems}, 35:10078--10093.

\bibitem[Wei et~al., 2022]{wei2022chain}
Wei, J., Wang, X., Schuurmans, D., Bosma, M., Xia, F., Chi, E., Le, Q.~V., Zhou, D., et~al. (2022).
\newblock Chain-of-thought prompting elicits reasoning in large language models.
\newblock {\em Advances in neural information processing systems}, 35:24824--24837.

\bibitem[Zhang et~al., 2023]{zhang2023omni}
Zhang, J., Lehman, J., Stanley, K., and Clune, J. (2023).
\newblock Omni: Open-endedness via models of human notions of interestingness.
\newblock {\em arXiv preprint arXiv:2306.01711}.

\end{thebibliography}
